\documentclass[11pt,a4paper]{article}
\usepackage[hyperref]{acl2018}
\usepackage{times}
\usepackage{latexsym}
\usepackage[english]{babel}
\usepackage{url}
\usepackage{natbib}
\aclfinalcopy 
\usepackage{graphicx}
\usepackage{bm}
\usepackage{multirow}
\usepackage{pifont}
\usepackage{amsmath} 
\usepackage{amsthm}
\usepackage{algorithm}
\usepackage{algpseudocode}
\usepackage{varwidth}

\newtheorem{theorem}{Theorem}

\algnewcommand{\LeftComment}[1]{\Statex \(\triangleright\) #1}
\theoremstyle{definition}

\usepackage[inline]{enumitem}
\usepackage[toc,page]{appendix}
\usepackage{caption}
\usepackage[export]{adjustbox}
\usepackage[para]{footmisc}

\title{A Multi-task Learning Approach for Named Entity Recognition using Local Detection}
  
  \author{Nargiza Nosirova, Mingbin Xu, Hui Jiang \\
  Department of Electrical Engineering and Computer Science \\
  Lassonde School of Engineering, York University \\
  4700 Keele Street, Toronto, Ontario, Canada \\
  {\tt \{nana, xmb, hj\}@cse.yorku.ca}
 }

\date{}

\begin{document}
\maketitle
\begin{abstract}
Named entity recognition (NER) systems that perform well require task-related and manually annotated datasets. However, they are expensive to develop, and are thus limited in size. As there already exists a large number of NER datasets that share a certain degree of relationship but differ in content, it is important to explore the question of whether such datasets can be combined as a simple method for improving NER performance. To investigate this, we developed a novel locally detecting multi-task model using FFNNs. The model relies on encoding variable-length sequences of words into theoretically lossless and unique fixed-size representations.  We applied this method to several well-known NER tasks and compared the results of our model to baseline models as well as other published results. As a result, we observed competitive performance in nearly all of the tasks.
\end{abstract}

\section{Introduction}
Named entity recognition aims to solve the problem of detecting proper nouns in a text and categorizing them into different types of entities. Such information is useful for higher-level NLP applications such as summarization and question answering \cite{aramaki, ravichandran}.
NER systems have been originally built by applying hand-crafted features and other external resources to achieve good results \cite{ratinov}.
In the recent years, researchers have turned to neural network architectures. For example, \newcite{collobert2011} introduced a neural network model that learns important features from word embeddings, thus requiring little feature engineering. However, in his use of FFNNs, the context used around a word is restricted to a fixed-size window. This bears the risk of losing potentially relevant information between words that are far apart. Recently, \newcite{xu} proposed a local detection approach for NER by making use of a technique that can encode any variable-length sequence of words into a theoretically lossless and unique fixed-size representation. This technique, called the Fixed-size ordinally forgetting encoding (FOFE), has the ability to capture immediate dependencies within the sentence, and thus using the encodings as features partly overcomes the limitations of FFNNs. Using FOFE features practically eliminate any need for feature engineering. Furthermore, it is known that FFNNs are universal approximators, and its advantages over RNNs include easier tuning, faster training times, and a simpler implementation. Therefore, since the main drawbacks of FFNNs are resolved by FOFE features, they are acceptance for recognition. 

Meanwhile, learning many associated tasks in parallel has been shown to improve performance compared to learning each task separately \cite{bakker, caruana}. One of the more popular MTL approaches is hard-parameter sharing, which has the advantage of reducing the chance of over-fitting {baxter, 1997} while also being simple to implement. Generally, MTL is applied by using auxiliary tasks that are similar to the main task. For example, \newcite{martinez} uses auxiliary tasks such as Part-of-speech (POS) and chunking for main tasks such as NER. 

Our main contribution lies in combining these two proposed models for the NER task. We investigate how hard parameter sharing can be used to improve NER models, while also further exploring the idea of using auxiliary NER tasks to boost the performance of a main NER task. 
In this paper, we propose a novel multi-task FOFE-based FFNN model with the aim of generalizing the underlying distributions of the named entities in the data. We report our experimental results on several popular NER tasks. Our method has yielded competitive results and improved performance in comparison to the baselines in all tasks. 

This paper is organized as follows. Section 2 introduces the FOFE technique and shows how its encoding is lossless and unique. Section 3 outlines how we use FOFE to extract features for the NER task. Section 4 gives an overview of our model, including the MTL technique used. From Section 5 onwards, we outline the experimental setup, present our results and state our conclusion.  

\begin{figure}[t]
\centering
\includegraphics[width=0.35\textwidth, center]{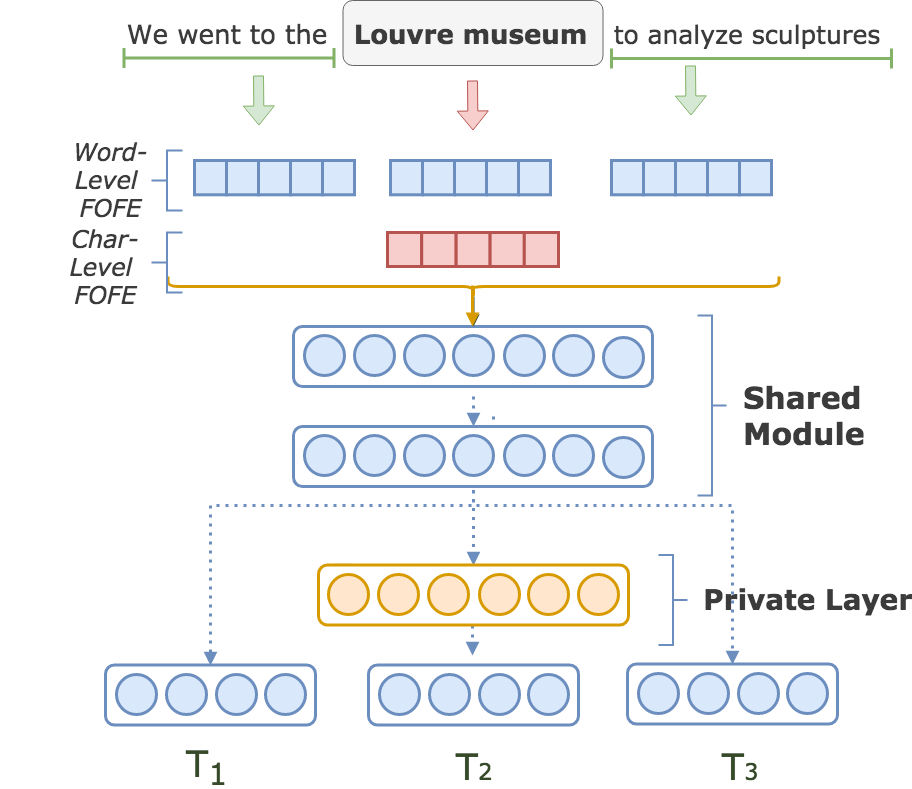}
\caption{Illustration of an example network structure for our MTL model using FOFE codes. The window currently examines the fragment {\it Louvre Museum}.}
\label{Fig:FOFE-NER-diagram}
\end{figure}

\section{Fixed-Size Ordinally Forgetting Encoding (FOFE) }
 Consider a vocabulary $V$, where each word can be represented by a 1-of-$|V|$ one-hot vector. Let $S = w_{1} \cdots  w_{N}$ denote a sequence of $N$ words from $V$, and denote $\bm{e_n}$ to be the one-hot vector of the $n$-{th} word in $S$, where $1 \leq n \leq N$.  Assuming  $\bm{z_0} = \mathbf{0}$,  the FOFE code $\bm{z_n}$ of the sequence from word $w_{1}$ to $w_{n}$ is as follows:
\begin{equation*}
\bm{z_n} = \alpha \cdot \bm{z_{n - 1}} + \bm{e_n}  \label{eq_FOFE_formula}
\end{equation*}
where $\alpha$ is a constant forgetting factor. Hence, $\bm{z_n}$ can be viewed as a fixed-size representation of the subsequence $\{ w_{1}, w_{2}, \cdots, w_{n} \}$.
Following the theoretical properties presented by \newcite{zhanghui} in Appendix A, we see that FOFE can uniquely and losslessly encode any sequence of variable length into a fixed-size representation.
\section{Extracting Features using FOFE }
\subparagraph{Word-level Features}
We extract the bag-of-words of the focus token as well as the FOFE encoding of its left and right contexts. All of the word features are computed in both case sensitive and case insensitive forms.  The FOFE encodings are further projected to lower-dimensional dense vectors using projection matrices for both the case sensitive and insensitive forms. 
Those matrices are initialized using word embeddings pre-trained with \textit{word2vec} \cite{mikolov}, and are tuned during training. 
\subparagraph{Character-level Features}
Based on a pre-defined set of all possible characters, we view the focus token as a case-sensitive character sequence and encode it using FOFE from left to right, as well as right-to-left. We then project the character encodings using a trainable character embedding matrix.  For a fair comparison, we also use character CNNs to generate additional character-level features \cite{kim}.
\section{A MTL approach for NER using FOFE}
Consider $k$ learning tasks $\left \{T_i \right \}_{i=1}^{k}$, where each task $T_{i}$ is associated with an input-output pair of sequences $(x_{1:n}, y^{i}_{1:n})$, where $x_{j} \in W$ and $y^{i}_{j} \in Y_{i}$. The input set $W$ is shared by all tasks, whereas the output sets $Y_{i}$ are reserved to a single corresponding task. At each training step, we randomly choose a task $T_{i}$ and training sample $(x_{1:n}, y^{i}_{1:n}) \in T_{i}$. We forward pass the training sample through the shared layers to predict the labels $\hat{y}^{i}_{j}$, calculate the loss based on the true labels $y^{i}_{j}$ and backpropagate for parameter update. The training sample of task $T_{i}$ is eventually fed into its corresponding task-specific softmax layer for classification, however the hidden layers are shared by all tasks. Additionally, we may attribute additional private hidden layers to  $T_{i}$, located between the shared layer and softmax layer.  If $T_{i}$ is a main task, the private layers can be useful for personalizing the learning of the task, since some of information contained in the training signals distinct to the task may be swamped in the shared layers by the auxiliary tasks' training signals. if  $T_{i}$ is an auxiliary task, it would enable the shared layers to focus on representing information pertinent to the other tasks, while keeping its distinct signals in its private layer. This is especially useful if $T_{i}$ has a large data size compared to the main task.

Figure \ref{Fig:FOFE-NER-diagram} represents an instance of the model. The character and word features extracted using FOFE are concatenated to form the input to the model. Each of the hidden layers are fully-connected. The model has many outputs, corresponding to the number of tasks trained for the specific instance.

\subparagraph*{Training} 
We use categorical cross entropy as our objective function. Training is executed by using mini-batch SGD with momentum of 0.9 \cite{bottou} and learning rates are exponentially decayed by a factor of 1/16 if dev-performance drops compared to the last run. We apply dropout \cite{srivastava} to all layers with a value of 0.5. We set all the forgetting factors for words to $\alpha_{w} = 0.5$, and to $\alpha_{c} = 0.8$ for characters. The layers are initialized based on a uniform distribution following \newcite{glorot}. 
All of the layers consist of ReLUs \cite{nairhinton}, and the probability of an output class is modelled by a softmax function. We follow the same post-processing and decoding steps for named entities as the ones outlined in \newcite{xu}. We performed grid search and selected the hyper-parameters over the main task's development set, with early stopping.  For detailed hyper-parameter settings, please go to Appendix A. 
\section{Experimental setup}
We use the following multi-lingual benchmark tasks: CoNLL-2003 ENG\footnote{ \cite{tjong2003}}, CoNLL-2002 SPA\footnote{ \cite{tjong2002}},  OntoNotes 5.0 ENG and ZH \footnote{\cite{pradhan}}, the KBP 2016 \footnote{\cite{kbp2016}} trilingual task and DEFT Light ERE. 
For each main task, we consider the following systems:
\begin{enumerate*}[label=(\roman*)]
	\item Baseline model, trained without any auxiliary task.
	\item A system involving a combination of auxiliary tasks along with the main task.
\end{enumerate*}
\begin{table*}[!htbp]
	\centering
	\scalebox{0.6}{
	\begin{tabular}{|l|lll|lll|lll|}
		\hline
		 & \multicolumn{3}{c|}{Single-task model} & \multicolumn{3}{c|}{Multi-task model} & \multicolumn{3}{c|}{Single-task 5-fold} \\ \hline
		 \multirow{3}{*}{LANG}  &
			\multicolumn{3}{c|}{\newcite{xu}} & \multicolumn{3}{c|}{This work} & \multicolumn{3}{c|}{2016 Best} \\
		 \ & P & R & F1 & P & R & F1 & P & R & F1 \\
		\hline \hline
		  	ENG & 0.836 & 0.680 & {\bf 0.750} & 0.800 & 0.742 & {\bf 0.770} & 0.846 & 0.710 & {\bf 0.772} \\
		   CMN & 0.789 & 0.625 & 0.698 & 0.766 & 0.673 & {\bf 0.717} & 0.789 & 0.737 & 0.762 \\
		   SPA & 0.835 & 0.602 & 0.700 & 0.869 & 0.618 & {\bf 0.722} & 0.839 & 0.656 & 0.736 \\
		   ALL & 0.819 & 0.639 & 0.718 & 0.806 & 0.676 & {\bf 0.738} & 0.802 & 0.704 & {\bf 0.756} \\
		\hline
	\end{tabular} }
	\caption{Comparison of the MTL models to \newcite{xu} and the best system for KBP 2016 task.}
	\label{tbl:kbp2016}
\end{table*}
\begin{table}
	\centering
	\scalebox{0.7}{
	\begin{tabular}{ | l l |}\hline
		\textbf{Model} & \textbf{ F1 (\%)} \\ \hline \hline
		\newcite{collobert2011} & $89.59$  \\
		\newcite{huang} & $90.10$ \\
		\newcite{strubell} & $90.54$ ( $\pm 0.18$ ) \\
		\newcite{yang} & $\mathbf{90.94}$ \\ \hline
		\newcite{luo} & $91.2$ \\
		\newcite{lample}  & $90.94$ \\
		\newcite{chiu} & $\mathbf{91.62}$ ($ \pm 0.33 $) \\ \hline
		\newcite{xu} & $90.71$ \\
		Our MTL model & $\mathbf{90.91}$   \\ \hline
	\end{tabular} }
	\caption{A comparison with baseline and top published results on CoNLL-2003 ENG eval. The three sections, in order, are models: trained with training set only, trained with both training and dev set, our baseline and model.}
	\label{tbl:conll-results-eng}
\end{table}
\begin{table}
	\centering
	\scalebox{0.7}{ 
	\begin{tabular}{ | l l |  } \hline
		\textbf{Model} & \textbf{ F1 (\%)} \\ \hline \hline
		\newcite{strubell} & $\mathbf{86.84}$ ($\pm 0.19$) \\
		\newcite{chiu} & $\mathbf{86.28}$ ($\pm 0.26$) \\
		\newcite{durrett} & $84.04$ \\ \hline
		\newcite{xu} & $85.88$ \\
		Our MTL model & $\mathbf{86.06}$ \\ \hline
	\end{tabular} }
	\caption{A comparison with baseline and state-of-the-art results on OntoNotes ENG NER Eval.}
	\label{tbl:ontonotes-results-eng}
\end{table}
\begin{table}
	\centering
	\scalebox{0.7}{
	\begin{tabular}{ | l l | } \hline
		\textbf{Model} & \textbf{ F1 (\%)} \\ \hline \hline
		\newcite{che} & $69.82$ \\
		\newcite{pappu} & $67.2$ \\ \hline
		\newcite{xu} & $71.83$ \\
		Our MTL model & $\mathbf{72.12}$ \\\hline
	\end{tabular} }
	\caption{A comparison with the baseline and other published results on OntoNotes Chinese NER Eval.}
	\label{tbl:ontonotes-results-cmn}
\end{table}
\begin{table}
\centering
	\scalebox{0.7}{
	\begin{tabular}{ | l l |  } \hline
		\textbf{Model} & \textbf{ F1 (\%)} \\ \hline \hline
		\newcite{santos} & $82.21$ \\
		\newcite{gillick} & $82.95$ \\
		\newcite{lample} & $\mathbf{85.75}$ \\
		\newcite{yang} & $\mathbf{85.77}$ \\ \hline
		\newcite{xu} & $83.22$ \\
		Our MTL model & $\mathbf{84.14}$ \\ \hline
	\end{tabular} }
	\caption{Results and comparison with the baseline and state-of-the-art results on CoNLL-2002 SPA NER Eval.}
	\label{tbl:conll-results-spa}
\end{table}
\subparagraph{Main and Auxiliary Tasks}
We group the tasks used together by language: 
\begin{enumerate*}[label=(\roman*)]  
\item ENG models: The CoNLL-2003, OntoNotes 5.0 and KBP 2016 ENG.
\item SPA models:  CoNLL-2002,  KBP 2016 SPA and Light ERE.
\item ZH models:  OntoNotes 5.0, KBP 2016 ZH and Light ERE.
\end{enumerate*}
Each task within the group is used as a main task in a separate experiment, and the rest of the tasks in the group are used as auxiliary tasks. The only exception is Light ERE, which is only auxiliary in all experiments.
Additionally, we make use an in-house dataset which consists of 10k ENG and ZH documents labelled manually following the KBP 2016 format. Since KBP 2016 does not contain any train and development data, we employ our in-house data as such with a 90:10 split. We also utilize the KBP 2015 dataset as additional data for training. For the CoNLL-2003 task, we use cased and uncased word embeddings of size $256$ trained on the Reuters RCV1 corpus. The remaining tasks have cased and uncased word embeddings of size $256$ trained using the English \footnote{\cite{parker2011english}}, Spanish \footnote{\cite{mendonca2009spanish}}  and Chinese \footnote{\cite{graff2005chinese}} Gigaword for the corresponding models evaluated in that language. Detailed info about the tasks can be found in Appendix A. 
\subparagraph*{Baselines}
Our baseline models are from \newcite{xu}. We use the author's findings for CoNLL-2003 and KBP 2016, and apply the implementation\footnote{\url{https://github.com/xmb-cipher/fofe-ner}} released by the author to train the model with OntoNotes 5.0 and CoNLL-2002 tasks.
\section{Results and Discussion}
Overall, the MTL models yield better performance over baselines for all of the tasks. The task that has most benefited from MTL is the KBP 2016 trilingual task, whose results are summarized in Table \ref{tbl:kbp2016}. \newcite{xu} and the best KBP 2016 system are single-task models and the latter used 5-fold cross-validation. All the KBP 2016 results have been generated using the official evaluator. 
Table \ref{tbl:summary-kbp2016-eng} summarizes the results obtained with MTL for the KBP 2016 ENG task. The first MTL model is trained by only using KBP 2015 data for the main task, which results in an $F_1$ score of 0.739, compared to only 0.697 in \newcite{xu}. We see that the gains experienced by MTL are more significant when the main task's training data is smaller. 
In Table \ref{tbl:conll-results-eng}, we compare our best MTL model for CoNLL-2003 with baseline and state-of-the-art results. Our model makes use of the task's training data only additionally to the auxiliary task training data. Compared to the models that are trained without the dev-set, our proposed model only comes second to \newcite{yang}. The OntoNotes 5.0 English and Chinese results are presented in Tables \ref{tbl:ontonotes-results-eng} and \ref{tbl:ontonotes-results-cmn}, and the CoNLL-2002 SPA results in Table \ref{tbl:conll-results-spa}. For ENG OntoNotes, we observe substantial gains over the baseline models, and are competitive with the top results. We should mention that we do not use any hand-crafted features. For the Chinese OntoNotes task, we found two non-neural method results and have exceeded them with both our baseline and MTL models. We have been unable to find prior published neural-based results, and thus cannot say with certainty whether we achieved state-of-the-art results. 
\begin{table}[!htbp]
	\centering
	\scalebox{0.7}{
	\begin{tabular}{|c|c|c|c|c|}
		\hline
			  Main task training data & MTL model & \newcite{xu} \\ \hline \hline
			KBP 2015 & {\bf 0.739} & {\bf 0.697} \\
			KBP 2015 + in-house &  0.770 & 0.750\\
      \hline
	\end{tabular} } 
	\caption{Results for KBP 2016 ENG with two different main task training set combinations.}
	\label{tbl:summary-kbp2016-eng}	
\end{table}
\section{Related Work}
\subparagraph{NER}
Recently, methods involving deep learning have been very successful in many NLP projects. Due to the limitations of FFNNs, more powerful neural networks, such as recurrent neural networks (RNNs) have been used. Many studies have used bidirectional Long Short-Term memory (B-LSTM) architecture along with CRF \cite{luo, huang}, and report convincing NER results. As for character-level modelling, studies have turned to convolutional neural networks (CNNs). For instance, \newcite{santos} have employed CNNs to extract character-level features for Spanish and Portuguese, and obtained successful results. 
\subparagraph{MTL}
Much of the work done in MTL has been initiated by \newcite{caruana}. His techniques have been used and confirmed in many studies \cite{maurer, andozhang}. The success of MTL has been associated with label entropy, regularizers, training size and many other aspects \cite{martinez, bingel}.  For example, \newcite{collobert2008} use MTL in a unified model to train multiple core NLP tasks: NER, Part-of-Speech, chunking and semantic role labeling with neural networks. They show that MTL improves generality among the shared tasks. \newcite{liu} used MTL for information retrieval and semantic classification by training a model for both tasks which has shared and private layers. Their method exceeded performance of strong baselines for tasks such as query classification and web search.
\section{Conclusion}
In this paper, we investigated the benefit of multi-task learning combined with local detection (FOFE) as a possible solution for improving performance on various NER tasks. We applied this method to several well-known NER tasks and observed competitive results, without using any external resources or hand-crafted features.

\appendix
\section{Supplemental Material}
\label{sec:supplemental}

\subsection{FOFE Theorems}
With $\alpha$ being the constant forgetting factor, the theoretical properties that show FOFE code uniqueness are as follows:
\begin{theorem}
If the forgetting factor $\alpha$ satisfies  $0 < \alpha \le 0.5$, FOFE is unique for any countable vocabulary $V$ and any finite value $N$ .
\end{theorem}

\begin{theorem}
For $0.5 < \alpha < 1$, given any finite value $N$ and any countable vocabulary $V$, FOFE is almost unique everywhere, except only a finite set of countable choices of $\alpha$. 
\end{theorem}

When $0.5 < \alpha < 1$, uniqueness is not guaranteed. However, the odds of ending up with such scenarios is small. Furthermore, it is rare to have a word reappear many times within a near context. Thus, we can say that FOFE can uniquely encode any sequence of variable length, providing a fixed-size lossless representation for any sequence. The proof for those theorems can be found in \newcite{zhanghui}.
\subsection{Data description}
\indent
\textit{CoNLL-2003:} The CoNLL-2003 dataset consists of newswire data originated from the Reuters RCV1 corpus. It is tagged with four entity types: person, location, organization and miscellaneous.
We only used the ENG documents in our experiments.

\textit{OntoNotes:} The OntoNotes dataset consists of text from sources such as broadcast conversation and news, newswire, telephone conversation, magazine and web text. The dataset was assembled by \newcite{pradhan} for the CoNLL-2012 shared task, who specifies a standard train, validation, and test split followed in our evaluation. It is tagged with eighteen entity types, some of which are: person, facility, organization, product, data, time, money, quantity and so forth. 

\textit{KBP 2016:} The KBP 2016 trilingual EDL task require the identification of entities (including nested) from a collection of text documents in three languages (ENG, ZH and SPA), and their classification to the following named and nominal entity types: person, geo-political entity, organization, location and facility. The dataset consists of recent news articles and discussion forums (non-parallel across languages). The KBP 2016 EDL task is an extension of the KBP 2015 task, except KBP 2015 does not contain any nominal types. We treat a named entity mention and its corresponding nominal mention as a single entity type and detect them together.

\textit{CoNLL-2002:} The CoNLL-2002 named entity data contains files covering both Spanish and Dutch, where each language has training, validation and evaluation files. Similarly to CoNLL-2003, It is tagged with four entity types: person, location, organization and miscellaneous. We mainly use the Spanish files for our Spanish NER model. 

\textit{Light ERE:} The DEFT Light ERE dataset consists of discussion forum and newswire documents tagged with five types of named entities: person, title, organization, geopolitical entities and location. 

\textit{In-house dataset:} Our in-house dataset consists of 10k English and Chinese documents that are labelled manually following the KBP 2016 dataset. 

\subsection{Training details}
\subparagraph*{Hyper-parameters}

\begin{itemize}
\item \textit{CoNLL-2003 ENG:} The model has two hidden layers in the shared module and contains a private module for the OntoNotes task with one hidden layer. The hidden layers in the shared module contain 700 units, while the one in the private layer has 512 units. Training is done by mini-batch of size 256. The learning rate is set to 0.128. We used case-sensitive and insensitive word embeddings of 256 dimensions trained using Reuters RCV1, and randomly initialized character embeddings of dimension 64. The official training, development and test set partition is used. 

\item \textit{OntoNotes ENG:} The multi-task model setup for this dataset is the same as the one for CoNLL-2003, except we use a learning rate of 0.064 and mini-batch of size 128. We follow the split dictated by \newcite{pradhan}. Also, the word embeddings are derived from the English Gigaword instead \cite{parker2011english}. Baseline: The baseline model is an FFNN with fully-connected ReLU activation layers that lead to a single output layer with softmax activation. It contains two hidden layers of size 512. The learning rate is set to 0.128, and the mini-batch size is 256.

\item \textit{KBP 2016:} For each language, we set up three models that are trained and evaluated independently. We use three sets of word embeddings of 256 dimensions from the English, Spanish \cite{mendonca2009spanish}  and Chinese \cite{graff2005chinese} Gigaword. As specified in \newcite{xu}, Chinese is labelled at character level only. Here is an overview for each of the models:
\begin{enumerate}
	\item English and Chinese: Similar to CoNLL-2003, however the private module is instead dedicated to the KBP 2016 task. The learning rate is set to 0.064 with a mini-batch size of 128.
	\item Spanish: Contains a shared module only, with two hidden layers of size 612. The learning rate is set to 0.128, with a mini-batch size of 128. 
\end{enumerate}

\item \textit{OntoNotes ZH:} The multi-task model set up for this dataset is the same as the one for the Chinese KBP model, with instead a private module for the OntoNotes task, two shared hidden layers of size 712 and a private hidden layer of size 512.

\item \textit{CoNLL-2002:} Contains a shared module only, with two hidden layers of size 612. The learning rate is set to 0.256, with a mini-batch size of 128.
Baseline: We set up the CoNLL-2002 baseline model in the same way as the OntoNotes baseline model, with hidden layers of size 412. 
\end{itemize}
 \subparagraph*{Effect of auxiliary training  data size} We ran all of our systems by gradually increasing the size of the auxiliary tasks training data in 20\% increments, while keeping the size of the main task constant. We did not observe any significant improvements over the baseline for any combination. We noticed that adding private hidden layers to some of the auxiliary tasks instead brought more benefit to the model performance. 

\bibliography{acl2018}

\begin{thebibliography}{38}
\expandafter\ifx\csname natexlab\endcsname\relax\def\natexlab#1{#1}\fi

\bibitem[{Ando and Zhang(2005)}]{andozhang}
Rie~Kubota Ando and Tong Zhang. 2005.
\newblock \href {http://dl.acm.org/citation.cfm?id=1046920.1194905} {A
  framework for learning predictive structures from multiple tasks and
  unlabeled data}.
\newblock \emph{J. Mach. Learn. Res.}, 6:1817--1853.

\bibitem[{Aramaki et~al.(2009)Aramaki, Miura, Tonoike, Ohkuma, Mashuichi, and
  Ohe}]{aramaki}
Eiji Aramaki, Yasuhide Miura, Masatsugu Tonoike, Tomoko Ohkuma, Hiroshi
  Mashuichi, and Kazuhiko Ohe. 2009.
\newblock \href {http://www.aclweb.org/anthology/W09-1324} {Text2table: Medical
  text summarization system based on named entity recognition and modality
  identification}.
\newblock In \emph{Proceedings of the BioNLP 2009 Workshop}, pages 185--192.
  Association for Computational Linguistics.

\bibitem[{Bakker and Heskes(2003)}]{bakker}
Bart Bakker and Tom Heskes. 2003.
\newblock \href {https://doi.org/10.1162/153244304322765658} {Task clustering
  and gating for bayesian multitask learning}.
\newblock \emph{J. Mach. Learn. Res.}, 4:83--99.

\bibitem[{Bingel and S{\o}gaard(2017)}]{bingel}
Joachim Bingel and Anders S{\o}gaard. 2017.
\newblock \href {http://aclweb.org/anthology/E17-2026} {Identifying beneficial
  task relations for multi-task learning in deep neural networks}.
\newblock In \emph{Proceedings of the 15th Conference of the European Chapter
  of the Association for Computational Linguistics: Volume 2, Short Papers},
  pages 164--169. Association for Computational Linguistics.

\bibitem[{Bottou(2010)}]{bottou}
L{\'e}on Bottou. 2010.
\newblock Large-scale machine learning with stochastic gradient descent.
\newblock In \emph{Proceedings of COMPSTAT'2010}, pages 177--186, Heidelberg.
  Physica-Verlag HD.

\bibitem[{Caruana(1997)}]{caruana}
Rich Caruana. 1997.
\newblock \href {https://doi.org/10.1023/A:1007379606734} {Multitask learning}.
\newblock \emph{Mach. Learn.}, 28(1):41--75.

\bibitem[{Che et~al.(2013)Che, Wang, Manning, and Liu}]{che}
Wanxiang Che, Mengqiu Wang, Christopher~D. Manning, and Ting Liu. 2013.
\newblock \href {http://www.aclweb.org/anthology/N13-1006} {Named entity
  recognition with bilingual constraints}.
\newblock In \emph{Proceedings of the 2013 Conference of the North American
  Chapter of the Association for Computational Linguistics: Human Language
  Technologies}, pages 52--62, Atlanta, Georgia. Association for Computational
  Linguistics.

\bibitem[{Chiu and Nichols(2016)}]{chiu}
Jason Chiu and Eric Nichols. 2016.
\newblock \href {https://transacl.org/ojs/index.php/tacl/article/view/792}
  {Named entity recognition with bidirectional lstm-cnns}.
\newblock \emph{Transactions of the Association for Computational Linguistics},
  4:357--370.

\bibitem[{Collobert and Weston(2008)}]{collobert2008}
Ronan Collobert and Jason Weston. 2008.
\newblock \href {https://doi.org/10.1145/1390156.1390177} {A unified
  architecture for natural language processing: Deep neural networks with
  multitask learning}.
\newblock In \emph{Proceedings of the 25th International Conference on Machine
  Learning}, ICML '08, pages 160--167, New York, NY, USA. ACM.

\bibitem[{Collobert et~al.(2011)Collobert, Weston, Bottou, Karlen, Kavukcuoglu,
  and Kuksa}]{collobert2011}
Ronan Collobert, Jason Weston, L{\'e}on Bottou, Michael Karlen, Koray
  Kavukcuoglu, and Pavel Kuksa. 2011.
\newblock \href {http://dl.acm.org/citation.cfm?id=1953048.2078186} {Natural
  language processing (almost) from scratch}.
\newblock \emph{J. Mach. Learn. Res.}, 12:2493--2537.

\bibitem[{Durrett and Klein(2014)}]{durrett}
Greg Durrett and Dan Klein. 2014.
\newblock \href {https://transacl.org/ojs/index.php/tacl/article/view/412} {A
  joint model for entity analysis: Coreference, typing, and linking}.
\newblock \emph{Transactions of the Association for Computational Linguistics},
  2:477--490.

\bibitem[{Gillick et~al.(2016)Gillick, Brunk, Vinyals, and
  Subramanya}]{gillick}
Dan Gillick, Cliff Brunk, Oriol Vinyals, and Amarnag Subramanya. 2016.
\newblock \href {http://www.aclweb.org/anthology/N16-1155} {Multilingual
  language processing from bytes}.
\newblock In \emph{Proceedings of the 2016 Conference of the North American
  Chapter of the Association for Computational Linguistics: Human Language
  Technologies}, pages 1296--1306, San Diego, California. Association for
  Computational Linguistics.

\bibitem[{Glorot et~al.(2011)Glorot, Bordes, and Bengio}]{glorot}
Xavier Glorot, Antoine Bordes, and Yoshua Bengio. 2011.
\newblock \href {http://proceedings.mlr.press/v15/glorot11a.html} {Deep sparse
  rectifier neural networks}.
\newblock In \emph{Proceedings of the Fourteenth International Conference on
  Artificial Intelligence and Statistics}, volume~15 of \emph{Proceedings of
  Machine Learning Research}, pages 315--323, Fort Lauderdale, FL, USA. PMLR.

\bibitem[{Graff and Chen(2005)}]{graff2005chinese}
David Graff and Ke~Chen. 2005.
\newblock Chinese gigaword.
\newblock \emph{LDC Catalog No.: LDC2003T09, ISBN}, 1:58563--58230.

\bibitem[{Huang et~al.(2015)Huang, Xu, and Yu}]{huang}
Zhiheng Huang, Wei Xu, and Kai Yu. 2015.
\newblock \href {http://arxiv.org/abs/1508.01991} {Bidirectional {LSTM-CRF}
  models for sequence tagging}.
\newblock \emph{CoRR}, abs/1508.01991.

\bibitem[{Ji and Nothman(2016)}]{kbp2016}
Heng Ji and Joel Nothman. 2016.
\newblock \href
  {https://tac.nist.gov/publications/2016/additional.papers/TAC2016.KBP_Entity_Discovery_and_Linking_overview.proceedings.pdf}
  {Overview of tac-kbp2016 tri-lingual edl and its impact on end-to-end
  cold-start kbp}.
\newblock In \emph{Proceedings of the Nineth Text Analysis Conference
  (TAC2016)}.

\bibitem[{Kim et~al.(2015)Kim, Jernite, Sontag, and Rush}]{kim}
Yoon Kim, Yacine Jernite, David Sontag, and Alexander~M. Rush. 2015.
\newblock \href {http://arxiv.org/abs/1508.06615} {Character-aware neural
  language models}.
\newblock \emph{CoRR}, abs/1508.06615.

\bibitem[{Lample et~al.(2016)Lample, Ballesteros, Subramanian, Kawakami, and
  Dyer}]{lample}
Guillaume Lample, Miguel Ballesteros, Sandeep Subramanian, Kazuya Kawakami, and
  Chris Dyer. 2016.
\newblock \href {https://doi.org/10.18653/v1/N16-1030} {Neural architectures
  for named entity recognition}.
\newblock In \emph{Proceedings of the 2016 Conference of the North American
  Chapter of the Association for Computational Linguistics: Human Language
  Technologies}, pages 260--270. Association for Computational Linguistics.

\bibitem[{Liu et~al.(2015)Liu, Gao, He, Deng, Duh, and Wang}]{liu}
Xiaodong Liu, Jianfeng Gao, Xiaodong He, Li~Deng, Kevin Duh, and Ye-Yi Wang.
  2015.
\newblock \href
  {http://dblp.uni-trier.de/db/conf/naacl/naacl2015.html#LiuGHDDW15}
  {Representation learning using multi-task deep neural networks for semantic
  classification and information retrieval.}
\newblock In \emph{HLT-NAACL}, pages 912--921. The Association for
  Computational Linguistics.

\bibitem[{Luo et~al.(2015)Luo, Huang, Lin, and Nie}]{luo}
Gang Luo, Xiaojiang Huang, Chin-Yew Lin, and Zaiqing Nie. 2015.
\newblock Joint entity recognition and disambiguation.
\newblock In \emph{Proceedings of the 2015 Conference on Empirical Methods in
  Natural Language Processing}, pages 879--888. Association for Computational
  Linguistics.

\bibitem[{Mart\'{i}nez~Alonso and Plank(2017)}]{martinez}
H\'{e}ctor Mart\'{i}nez~Alonso and Barbara Plank. 2017.
\newblock \href {http://www.aclweb.org/anthology/E17-1005} {When is multitask
  learning effective? semantic sequence prediction under varying data
  conditions}.
\newblock In \emph{Proceedings of the 15th Conference of the European Chapter
  of the Association for Computational Linguistics: Volume 1, Long Papers},
  pages 44--53, Valencia, Spain. Association for Computational Linguistics.

\bibitem[{Maurer et~al.(2016)Maurer, Pontil, and Romera-Paredes}]{maurer}
Andreas Maurer, Massimiliano Pontil, and Bernardino Romera-Paredes. 2016.
\newblock \href {http://dl.acm.org/citation.cfm?id=2946645.3007034} {The
  benefit of multitask representation learning}.
\newblock \emph{J. Mach. Learn. Res.}, 17(1):2853--2884.

\bibitem[{Mendonca et~al.(2009)Mendonca, Graff, and
  DiPersio}]{mendonca2009spanish}
Angelo Mendonca, David~Andrew Graff, and Denise DiPersio. 2009.
\newblock \emph{Spanish gigaword second edition}.
\newblock Linguistic Data Consortium.

\bibitem[{Mikolov et~al.(2013)Mikolov, Sutskever, Chen, Corrado, and
  Dean}]{mikolov}
Tomas Mikolov, Ilya Sutskever, Kai Chen, Greg Corrado, and Jeffrey Dean. 2013.
\newblock \href {http://dl.acm.org/citation.cfm?id=2999792.2999959}
  {Distributed representations of words and phrases and their
  compositionality}.
\newblock In \emph{Proceedings of the 26th International Conference on Neural
  Information Processing Systems - Volume 2}, NIPS'13, pages 3111--3119, USA.
  Curran Associates Inc.

\bibitem[{Nair and Hinton(2010)}]{nairhinton}
Vinod Nair and Geoffrey~E. Hinton. 2010.
\newblock \href {http://dl.acm.org/citation.cfm?id=3104322.3104425} {Rectified
  linear units improve restricted boltzmann machines}.
\newblock In \emph{Proceedings of the 27th International Conference on
  International Conference on Machine Learning}, ICML'10, pages 807--814, USA.
  Omnipress.

\bibitem[{Pappu et~al.(2017)Pappu, Blanco, Mehdad, Stent, and Thadani}]{pappu}
Aasish Pappu, Roi Blanco, Yashar Mehdad, Amanda Stent, and Kapil Thadani. 2017.
\newblock \href {https://doi.org/10.1145/3018661.3018724} {Lightweight
  multilingual entity extraction and linking}.
\newblock In \emph{Proceedings of the Tenth ACM International Conference on Web
  Search and Data Mining}, WSDM '17, pages 365--374, New York, NY, USA. ACM.

\bibitem[{Parker et~al.(2011)Parker, Graff, Kong, Chen, and
  Maeda}]{parker2011english}
Robert Parker, David Graff, Junbo Kong, Ke~Chen, and Kazuaki Maeda. 2011.
\newblock English gigaword.
\newblock \emph{Linguistic Data Consortium}.

\bibitem[{Pradhan et~al.(2013)Pradhan, Moschitti, Xue, Ng, Bj{ö}rkelund,
  Uryupina, Zhang, and Zhong}]{pradhan}
Sameer Pradhan, Alessandro Moschitti, Nianwen Xue, Hwee~Tou Ng, Anders
  Bj{ö}rkelund, Olga Uryupina, Yuchen Zhang, and Zhi Zhong. 2013.
\newblock \href {http://www.aclweb.org/anthology/W13-3516} {Towards robust
  linguistic analysis using ontonotes}.
\newblock In \emph{Proceedings of the Seventeenth Conference on Computational
  Natural Language Learning}, pages 143--152, Sofia, Bulgaria. Association for
  Computational Linguistics.

\bibitem[{Ratinov and Roth(2009)}]{ratinov}
Lev Ratinov and Dan Roth. 2009.
\newblock \href {http://dl.acm.org/citation.cfm?id=1596374.1596399} {Design
  challenges and misconceptions in named entity recognition}.
\newblock In \emph{Proceedings of the Thirteenth Conference on Computational
  Natural Language Learning}, CoNLL '09, pages 147--155, Stroudsburg, PA, USA.
  Association for Computational Linguistics.

\bibitem[{Ravichandran and Hovy(2002)}]{ravichandran}
Deepak Ravichandran and Eduard Hovy. 2002.
\newblock \href {https://doi.org/10.3115/1073083.1073092} {Learning surface
  text patterns for a question answering system}.
\newblock In \emph{Proceedings of the 40th Annual Meeting on Association for
  Computational Linguistics}, ACL '02, pages 41--47, Stroudsburg, PA, USA.
  Association for Computational Linguistics.

\bibitem[{dos Santos and Guimar{\~{a}}es(2015)}]{santos}
C{\'{\i}}cero~Nogueira dos Santos and Victor Guimar{\~{a}}es. 2015.
\newblock \href {http://arxiv.org/abs/1505.05008} {Boosting named entity
  recognition with neural character embeddings}.
\newblock \emph{CoRR}, abs/1505.05008.

\bibitem[{Srivastava et~al.(2014)Srivastava, Hinton, Krizhevsky, Sutskever, and
  Salakhutdinov}]{srivastava}
Nitish Srivastava, Geoffrey Hinton, Alex Krizhevsky, Ilya Sutskever, and Ruslan
  Salakhutdinov. 2014.
\newblock \href {http://dl.acm.org/citation.cfm?id=2627435.2670313} {Dropout: A
  simple way to prevent neural networks from overfitting}.
\newblock \emph{J. Mach. Learn. Res.}, 15(1):1929--1958.

\bibitem[{Strubell et~al.(2017)Strubell, Verga, Belanger, and
  McCallum}]{strubell}
Emma Strubell, Patrick Verga, David Belanger, and Andrew McCallum. 2017.
\newblock \href {https://aclanthology.info/papers/D17-1283/d17-1283} {Fast and
  accurate entity recognition with iterated dilated convolutions}.
\newblock In \emph{Proceedings of the 2017 Conference on Empirical Methods in
  Natural Language Processing, {EMNLP} 2017, Copenhagen, Denmark, September
  9-11, 2017}, pages 2670--2680.

\bibitem[{Tjong Kim~Sang(2002)}]{tjong2002}
Erik~F. Tjong Kim~Sang. 2002.
\newblock \href {https://doi.org/10.3115/1118853.1118877} {Introduction to the
  conll-2002 shared task: Language-independent named entity recognition}.
\newblock In \emph{Proceedings of the 6th Conference on Natural Language
  Learning - Volume 20}, COLING-02, pages 1--4, Stroudsburg, PA, USA.
  Association for Computational Linguistics.

\bibitem[{Tjong Kim~Sang and De~Meulder(2003)}]{tjong2003}
Erik~F. Tjong Kim~Sang and Fien De~Meulder. 2003.
\newblock \href {https://doi.org/10.3115/1119176.1119195} {Introduction to the
  conll-2003 shared task: Language-independent named entity recognition}.
\newblock In \emph{Proceedings of the Seventh Conference on Natural Language
  Learning at HLT-NAACL 2003 - Volume 4}, CONLL '03, pages 142--147,
  Stroudsburg, PA, USA. Association for Computational Linguistics.

\bibitem[{Xu et~al.(2017)Xu, Jiang, and Watcharawittayakul}]{xu}
Mingbin Xu, Hui Jiang, and Sedtawut Watcharawittayakul. 2017.
\newblock \href {http://aclweb.org/anthology/P17-1114} {A local detection
  approach for named entity recognition and mention detection}.
\newblock In \emph{Proceedings of the 55th Annual Meeting of the Association
  for Computational Linguistics (Volume 1: Long Papers)}, pages 1237--1247,
  Vancouver, Canada. Association for Computational Linguistics.

\bibitem[{Yang et~al.(2016)Yang, Salakhutdinov, and Cohen}]{yang}
Zhilin Yang, Ruslan Salakhutdinov, and William~W. Cohen. 2016.
\newblock \href {http://arxiv.org/abs/1603.06270} {Multi-task cross-lingual
  sequence tagging from scratch}.
\newblock \emph{CoRR}, abs/1603.06270.

\bibitem[{Zhang et~al.(2015)Zhang, Jiang, Xu, Hou, and Dai}]{zhanghui}
Shiliang Zhang, Hui Jiang, Mingbin Xu, Junfeng Hou, and Li{-}Rong Dai. 2015.
\newblock \href {http://arxiv.org/abs/1505.01504} {A fixed-size encoding method
  for variable-length sequences with its application to neural network language
  models}.
\newblock \emph{CoRR}, abs/1505.01504.

\end{thebibliography}


\begin{thebibliography}{6}
\expandafter\ifx\csname natexlab\endcsname\relax\def\natexlab#1{#1}\fi

\bibitem[{Graff and Chen(2005)}]{graff2005chinese}
David Graff and Ke~Chen. 2005.
\newblock Chinese gigaword.
\newblock \emph{LDC Catalog No.: LDC2003T09, ISBN}, 1:58563--58230.

\bibitem[{Mendonca et~al.(2009)Mendonca, Graff, and
  DiPersio}]{mendonca2009spanish}
Angelo Mendonca, David~Andrew Graff, and Denise DiPersio. 2009.
\newblock \emph{Spanish gigaword second edition}.
\newblock Linguistic Data Consortium.

\bibitem[{Parker et~al.(2011)Parker, Graff, Kong, Chen, and
  Maeda}]{parker2011english}
Robert Parker, David Graff, Junbo Kong, Ke~Chen, and Kazuaki Maeda. 2011.
\newblock English gigaword.
\newblock \emph{Linguistic Data Consortium}.

\bibitem[{Pradhan et~al.(2013)Pradhan, Moschitti, Xue, Ng, Bj{ö}rkelund,
  Uryupina, Zhang, and Zhong}]{pradhan}
Sameer Pradhan, Alessandro Moschitti, Nianwen Xue, Hwee~Tou Ng, Anders
  Bj{ö}rkelund, Olga Uryupina, Yuchen Zhang, and Zhi Zhong. 2013.
\newblock \href {http://www.aclweb.org/anthology/W13-3516} {Towards robust
  linguistic analysis using ontonotes}.
\newblock In \emph{Proceedings of the Seventeenth Conference on Computational
  Natural Language Learning}, pages 143--152, Sofia, Bulgaria. Association for
  Computational Linguistics.

\bibitem[{Xu et~al.(2017)Xu, Jiang, and Watcharawittayakul}]{xu}
Mingbin Xu, Hui Jiang, and Sedtawut Watcharawittayakul. 2017.
\newblock \href {http://aclweb.org/anthology/P17-1114} {A local detection
  approach for named entity recognition and mention detection}.
\newblock In \emph{Proceedings of the 55th Annual Meeting of the Association
  for Computational Linguistics (Volume 1: Long Papers)}, pages 1237--1247,
  Vancouver, Canada. Association for Computational Linguistics.

\bibitem[{Zhang et~al.(2015)Zhang, Jiang, Xu, Hou, and Dai}]{zhanghui}
Shiliang Zhang, Hui Jiang, Mingbin Xu, Junfeng Hou, and Li{-}Rong Dai. 2015.
\newblock \href {http://arxiv.org/abs/1505.01504} {A fixed-size encoding method
  for variable-length sequences with its application to neural network language
  models}.
\newblock \emph{CoRR}, abs/1505.01504.

\end{thebibliography}
\bibliographystyle{acl_natbib}

\end{document}